\documentclass[lettersize,journal]{IEEEtran}
\usepackage{amsmath,amsfonts}
\usepackage{algorithmic}
\usepackage{algorithm}
\usepackage{array}
\usepackage[caption=false,font=normalsize,labelfont=sf,textfont=sf]{subfig}
\usepackage{textcomp}
\usepackage{stfloats}
\usepackage{url}
\usepackage{verbatim}
\usepackage{graphicx}
\usepackage{cite}

\usepackage{xcolor}

\usepackage{multirow}

\usepackage[numbers,sort&compress]{natbib}

\usepackage[figuresleft]{rotating}

\begin{document}

\title{Milestones in Autonomous Driving and Intelligent Vehicles Part \uppercase\expandafter{\romannumeral2}: Perception and Planning}

\author{Long Chen, ~\IEEEmembership{Senior Member,~IEEE}, Siyu Teng, Bai Li, Xiaoxiang Na, Yuchen Li, Zixuan Li, Jinjun Wang, 

Dongpu Cao, ~\IEEEmembership{Senior Member,~IEEE}, Nanning Zheng,~\IEEEmembership{Fellow,~IEEE} and Fei-Yue Wang,~\IEEEmembership{Fellow,~IEEE} 
\thanks{Manuscript received Sep 30, 2022; revised May 30, 2023.
(Corresponding author: Fei-Yue Wang.)

This work is supported by the National Natural Science Foundation of China (62006256) and the Key Research and Development Program of Guangzhou (202007050002  202007050004).

Long Chen and Fei-Yue Wang are with the State Key Laboratory of Management and Control for Complex Systems, Institute of Automation, Chinese Academy of Sciences, Beijing, 100190, China, and Long Chen is also with Waytous Ltd. (e-mail: long.chen@ia.ac.cn; feiyue.wang@ia.ac.cn).

Siyu Teng and Yuchen Li are with BNU-HKBU United International College, Zhuhai, 519087, China and Hong Kong Baptist University, Kowloon, Hong Kong, 999077, China (e-mail: siyuteng@ieee.org; liyuchen2016@hotmail.com).

Bai Li is with the College of Mechanical and Vehicle Engineering, Hunan University (e-mail: libai@zju.edu.cn). 

Xiaoxiang Na is with the Department of Engineering, University of Cambridge (e-mail: xnhn2@cam.ac.uk).

Zixuan Li is with Waytous Ltd. (e-mail: lizixuan981258655@gmail.com).

Jinjun Wang and Nanning Zheng are with the College of Artificial Intelligence, Xi'an Jiaotong University (e-mail: jinjun@mail.xjtu.edu.cn; nnzheng@mail.xjtu.edu.cn).

Dongpu Cao is with the School of Mechanical Engineering, Tsinghua University (e-mail: dp\_cao2016@163.com).
}
}

\markboth{IEEE Transactions on Systems, Man, and Cybernetics: Systems,~Vol.~X, No.~X, X~X}
{Shell \MakeLowercase{\textit{et al.}}: A Sample Article Using IEEEtran.cls for IEEE Journals}


\maketitle

\begin{abstract}


Growing interest in autonomous driving (AD) and intelligent vehicles (IVs) is fueled by their promise for enhanced safety, efficiency, and economic benefits. While previous surveys have captured progress in this field, a comprehensive and forward-looking summary is needed. Our work fills this gap through three distinct articles. The first part, a "Survey of Surveys" (SoS), outlines the history, surveys, ethics, and future directions of AD and IV technologies. The second part, "Milestones in Autonomous Driving and Intelligent Vehicles Part I: Control, Computing System Design, Communication, HD Map, Testing, and Human Behaviors" delves into the development of control, computing system, communication, HD map, testing, and human behaviors in IVs. This part, the third part, reviews perception and planning in the context of IVs. Aiming to provide a comprehensive overview of the latest advancements in AD and IVs, this work caters to both newcomers and seasoned researchers. By integrating the SoS and Part I, we offer unique insights and strive to serve as a bridge between past achievements and future possibilities in this dynamic field.


\end{abstract}

\section{Introduction}



\IEEEPARstart{A}{utonomous} driving (AD) and intelligent vehicles (IVs) have recently attracted significant attention from academia as well as industry because of a range of potential benefits. Surveys on AD and IVs occupy an essential position in gathering research achievements, generalizing entire technology development, and forecasting future trends. However, a large majority of surveys only focus on specific tasks and lack systematic summaries and research directions in the future. As a result, they may have a negative impact on conducting research for abecedarians. Our work consists of 3 independent articles including a Survey of Surveys (SoS) \cite{SoS} and two surveys on crucial technologies of AD and IVs. Here is the third part (Part \uppercase\expandafter{\romannumeral2} of the survey) to systematically review the development of perception and planning. Combining with the SoS and the second part (Part \uppercase\expandafter{\romannumeral1} of the survey on control, system design, communication, High Definition map (HD map), testing, and human behaviors in IVs) \cite{Part1}, we expect that our work can be considered as a bridge between past and future for AD and IVs.

\begin{figure}
\centering  %
\includegraphics[width=9cm]{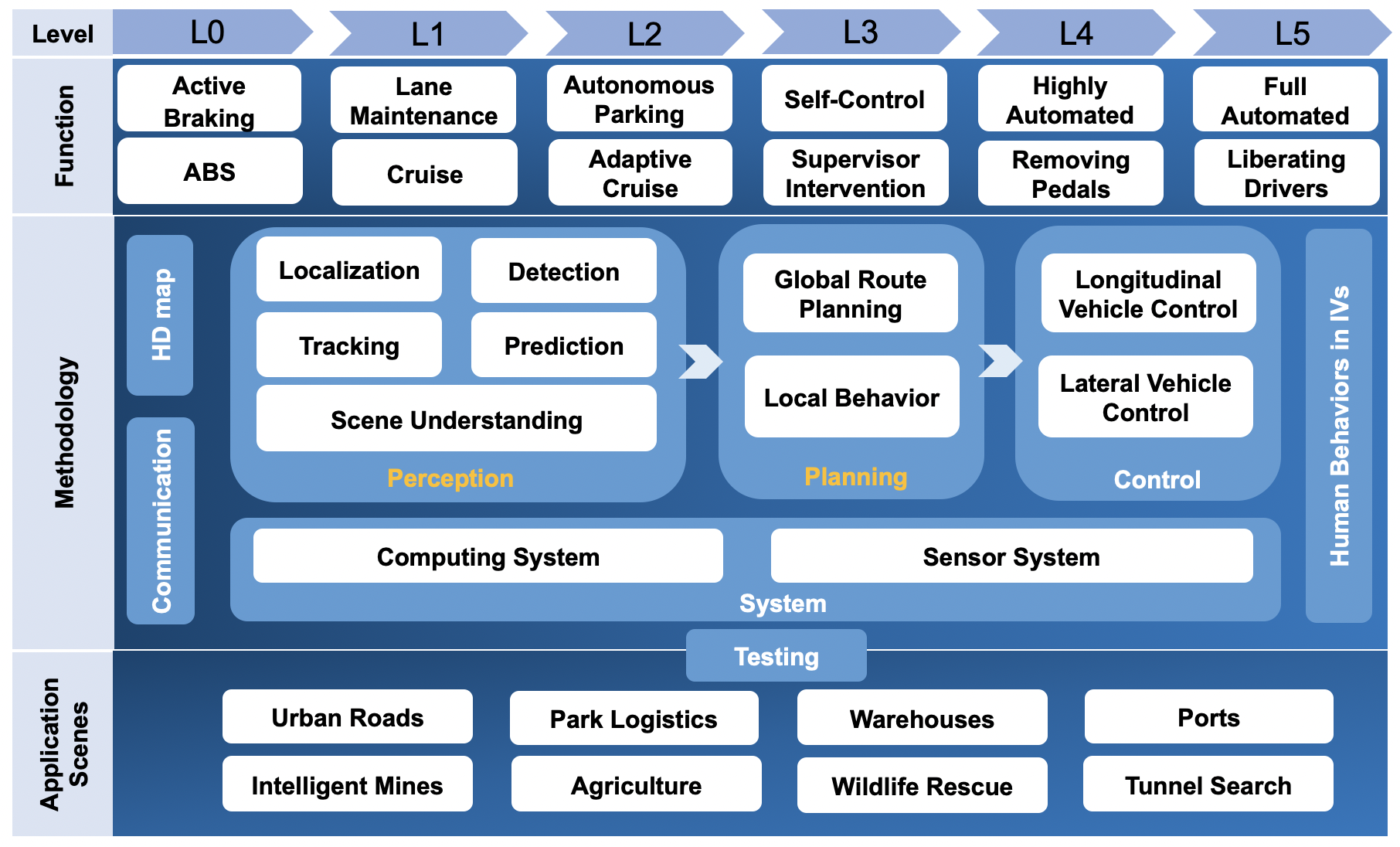}

\caption{The structure of autonomous driving with the function, methodology and application scenes}
\label{fig:structure}
\end{figure}

According to the different tasks in AD, we divide them into 8 sub-sections, perception, planning, control, system design, communication, HD map, testing, and human behaviors in IVs as Fig. \ref{fig:structure}. In Part \uppercase\expandafter{\romannumeral1}, we briefly introduce the function of each task and the intelligent levels for AD. Here, we describe classical applications in different AD scenes including urban roads, park logistics, warehouses, ports, intelligent mines, agriculture, wildlife rescue and tunnel search. It is more common for citizens to realize the AD in urban roads such as private IVs, AD taxis and buses. IVs in parts and ports require controllers to follow specific rules and achieve high efficiency. Warehouses and mines are classic closed scenes in indoor and outdoor environments. Modified IVs or called professional intelligent robots can be employed in wild to replace the human harbour in agricultural operations, wildlife rescue, tunnel search, etc. Indeed, AD and IVs could conduct a number of tasks in different scenes and play a crucial role in our daily life.

\begin{figure}
\centering  %
\includegraphics[width=8.5cm]{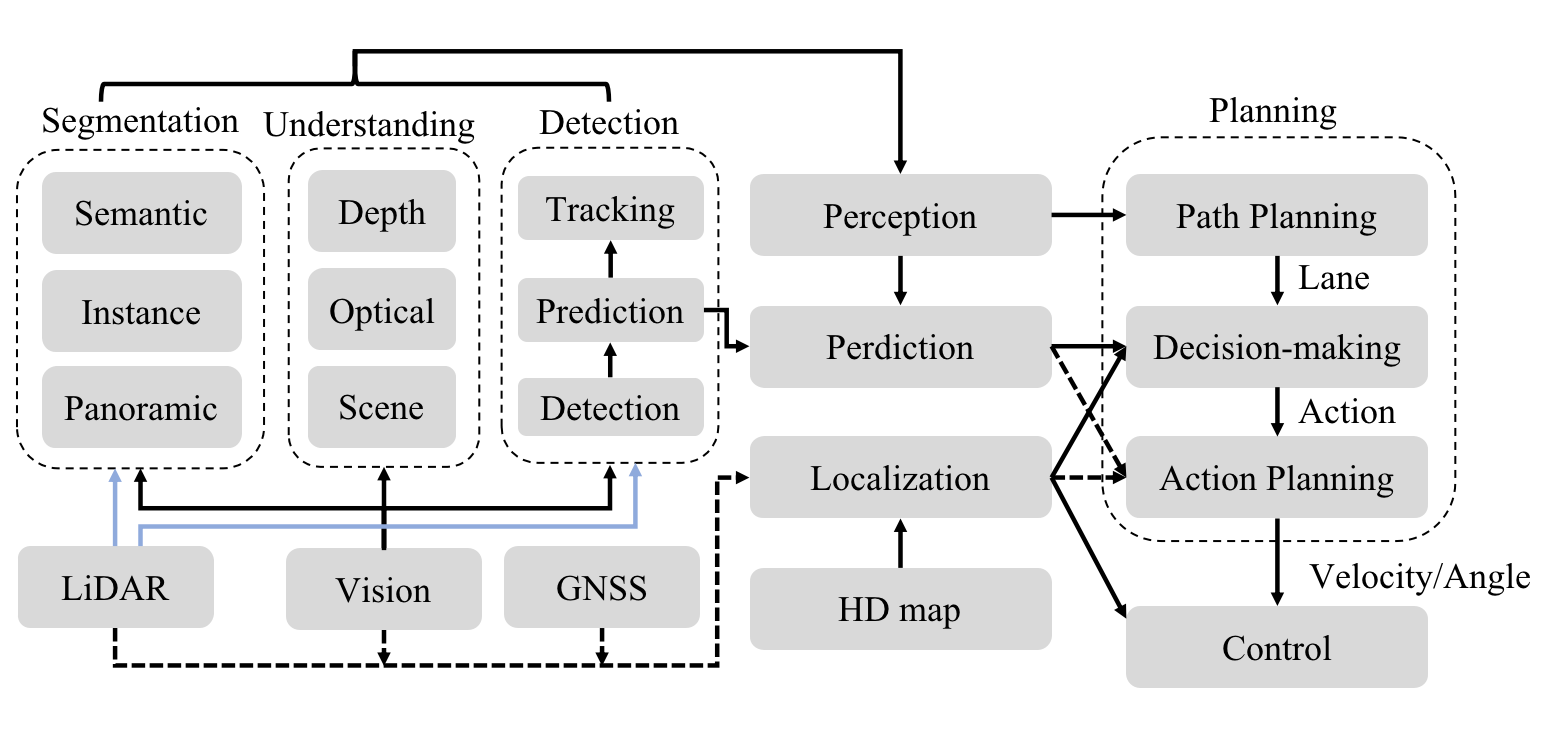}

\caption{The relationship of sub-tasks in perception, planning and the relationship of perception, planning and control}
\label{fig:relationship}
\end{figure}

In this paper, we consider $2$ sub-sections as independent chapters, and each of them includes task definition, functional divisions, novel ideas, and a detailed introduction to milestones of AD and IVs, and the relationship of perception, planning and control can be seen in Fig. \ref{fig:relationship}. The most important thing is that the research of them have rapidly developed for a decade and now entered a bottleneck period. We wish this article could be considered as a comprehensive summary for abecedarians and bring novel and diverse insights for researchers to make breakthroughs. We summarize three contributions of this article:

\quad 1. We provide a more systematic, comprehensive, and novel survey of crucial technology development with milestones on AD and IVs.

\quad 2. We introduce a number of deployment details, testing methods and unique insights throughout each technology section.

\quad 3. We conduct a systematic study that attempts to be a bridge between past and future on AD and IVs, and this article is the third part of our whole research (Part \uppercase\expandafter{\romannumeral2} for the survey).


\section{Perception}

\begin{figure}
\centering  
\includegraphics[width=8.5cm]{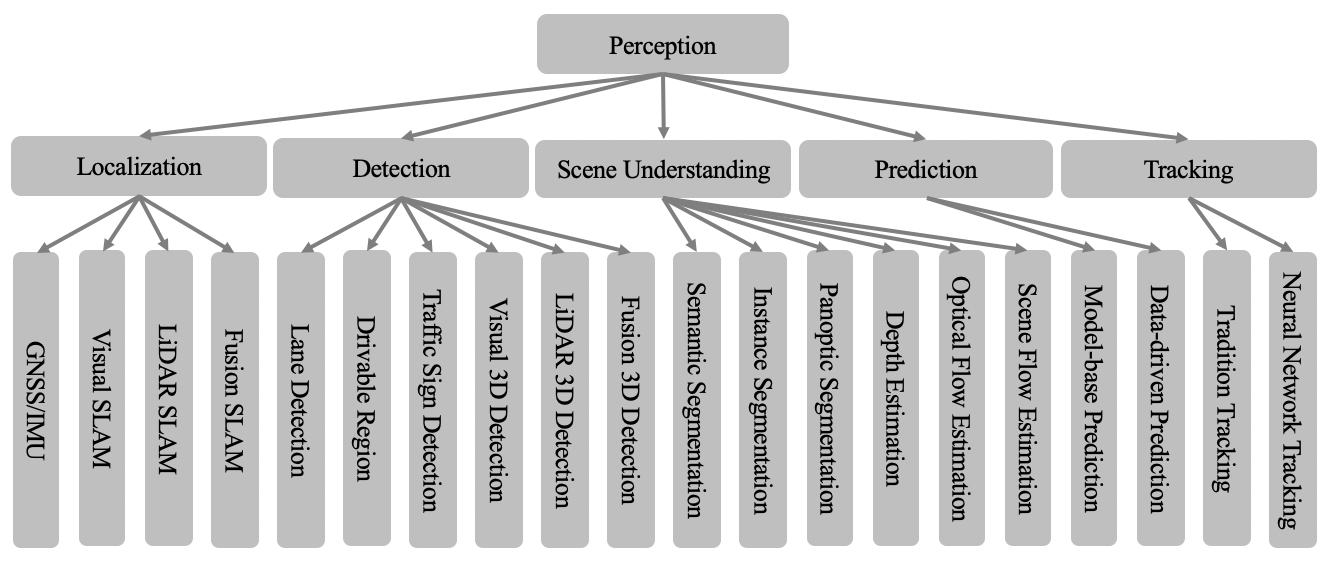}

\caption{The structure of perception methodology}
\label{fig:perception methodology}
\end{figure}

Perception is a fundamental module for AD. This module provides surrounding environmental information to the ego-vehicle. As can be seen in Fig. \ref{fig:perception methodology}, perception is divided into localization, object detection, scene understanding, target prediction, and tracking.

\subsection{\textbf{Localization}}
Localization is the technology for the driving platform to obtain its own position and attitude. It is an important prerequisite for the planning and control \cite{survey_loca_2}. Currently, localization strategies are divided into four categories: Global Navigation Satellite System (GNSS) and Inertial Measurement Unit (IMU), visual Simultaneously Localization and Mapping (SLAM), LiDAR SLAM, and fusion-based SLAM \cite{survey_loca_3}.

\subsubsection{GNSS and IMU}
The GNSS \cite{survey_loca_14} is a space-based radio navigation and localization system that can provide users with three-dimensional (3D) coordinates, velocity, and time information on the earth's surface. The IMU \cite{local_17} is commonly composed of three-axis accelerometers and gyroscopes (additional three-axis magnetometers for 9 Degree of Freedom (DOF) IMUs). By updating the localization information in low frequency from GNSS with dynamic states from IMUs, the platform could obtain the localization info in a high update frequency. Although the fusing the GNSS and IMU method is all-weather, the satellite signals will be interfered with by urban buildings \cite{local_36}.

\subsubsection{Visual SLAM} 
Visual SLAM adopts the change of frames from cameras to estimate the ego-vehicle motion and this type of algorithms is divided into three categories by sensors: the monocular, multiply views, and depth. Specifically, visual SLAM algorithms only require images as inputs that means the cost of the localization system is relatively cheap \cite{local_48}. However, they are dependent on abundant features and slight variation of illumination. In addition, optimization is a crucial module for visual localization system which updates each frame estimation after considering the global information and optimization methods include filter-based and graph-based \cite{local_24}.


There are two typical categories of visual SLAM from the perspective of feature extraction, key points \cite{ORB-SLAM, Stereo-DSO, S-PTAM, MonoSLAM} and optical flow methods \cite{LSD, DSO, SVO, DTAM}. Key points methods utilize points extraction approaches like SIFT, SURF, ORB, and descriptors to detect the same characteristics at different images and then compute relative motion among the frames. As points extraction approaches can extract crucial points stably and accurately, key points visual SLAM systems can offer significant benefits in structured roads and urban areas. However, the system may suffer difficulties when operating on an unstructured road or facing a flat white wall. Besides, earlier algorithms could not run in real-time and ignored most of the pixel information in the image. Optical flow methods assume that the photometric is invariant among the frames and attempt to estimate the camera motion by minimizing the photometric error on the images. This kind of method has several advantages as follows: 1) low computing overhead and high real-time performance; 2) weak dependence on key points; 3) considering whole pixels in the frames. However, due to the photometric assumption, it is sensitive for optical flow methods to the luminosity change between two images. visual SLAM systems also could be categorized into filter-based and optimization-based strategies from the optimization perspective, however, graph-based optimization methods have made a number of breakthroughs in accuracy and efficiency. Thus, researchers will continue to focus on the latter point in the future.

\subsubsection{LiDAR SLAM} 
Compared with the visual SLAM methods, LiDAR SLAM systems detect surrounding environments actively with accurate 3D information because of the LiDARs properties. Similar to visual systems, the LiDAR SLAM also could be categorized into 2D such as Gmapping, Cartographer, Karto, and 3D \cite{LOAM, LEGO-LOAM, SUMA, SegMap} methods by sensors or filter-based like Gmapping and optimization-based by the process of optimization. Gmapping adopts the particle filter approach and separates the localization and mapping processes. During the optimization, each particle is responsible for maintaining a map. LOAM \cite{LOAM} operates two parallel algorithms, one is to calculate the motion transformation between frames in a low frequency through point cloud matching methods, and the other attempts to construct a map and correct the odometry in a high frequency. Segmap \cite{SegMap} utilizes deep neural networks to extract semantic feature information, which could reduce the computational resource consumption, and solve the data compression problem in real-time for indoor intelligent robotics and IVs. SUMA \cite{SUMA} transfers the point clouds into 2D space and adopts an extended RGB-D SLAM structure to generate a local map. Besides, it maintains and updates the surfel map by Iterative Closest Point (ICP) matching method for point clouds. LiDAR SLAM systems have the advantages of high accuracy, achieving a dense map, and weak dependency on lightness. However, no semantic information and environmental disturbance are two main challenges for LiDAR SLAM systems. In addition, researchers have to spend lots of time and effort to maintain and repair LiDARs installed on the IVs.

\subsubsection{Fusion-based SLAM} 
In order to avoid the problems with failures in single sensor or low robustness, fusing multiple modalities data methods have been introduced by researchers including visual-inertial \cite{MSF-EKF, OKVIS, VINS-MONO}, LiDAR-inertial \cite{LIO-Mapping, LIOM, LINS, LIO-SAM}, visual-LiDAR inertial \cite{VLOAM, LIMO, VIL-SLAM} and other fusion, such as adding sonars \cite{Sonar-SLAM} or radars \cite{Radar-SLAM}, SLAM approaches. We found that fusion methods usually introduce IMU data with higher updating frequency to SLAM systems. Loose fusion methods \cite{LOAM, VLOAM} treat the external observation data from cameras or LIDARs and the internal motion data from IMUs as two independent modules, while the tight fusion approaches \cite{VINS-MONO, LIOM, LIMO, VIL-SLAM} design a unit optimization module to solve and fuse multiple modalities data. Former methods could be considered as extended visual or LiDAR SLAM systems and are friendly for researchers to deploy on testing platforms and IVs. However, to increase the Robustness and adaptability, the tight fusion strategies provide appropriate solutions including introducing bundle adjustment into the visual odometry system \cite{LIMO} and adopting association optimization \cite{VIL-SLAM}. In summary, fusion-based SLAM methods solve several difficulties for a single sensor but still introduce a few challenges for jointing systems such as calibration, synchronization, and complex processing. The advantages and disadvantages of different methods for localization are shown in TABLE \ref{table:control_table_1}.

\begin{table}[]
\caption{The Advantages and Disadvantages of Different Localization Methods}

\begin{tabular}{cll}
\hline
Method        & Advantages & Disadvantages                                                                                       \\\hline
GNSS & \begin{tabular}[c]{@{}l@{}}1.All-weather\\2.Easy configuration\\3.Non external sensors \end{tabular}           &\begin{tabular}[c]{@{}l@{}}1.Site requirements  \\2.GPS update frequency \\3.High-cost (high accuracy) \end{tabular}                                    \\\hline
Visual & \begin{tabular}[c]{@{}l@{}}1.Economy\\2.Rich semantic info \end{tabular}                               &\begin{tabular}[c]{@{}l@{}}1.Lightness effect \\2.Similar feature \\3.Low accuracy \end{tabular}                                                                     \\\hline
LiDAR    & \begin{tabular}[c]{@{}l@{}}1.High accuracy. \\2.Dense map\end{tabular}                 & \begin{tabular}[c]{@{}l@{}}1.Non semantic info\\2.Environmental disturbance \\3.High-cost\end{tabular}  \\\hline
Fusion            & \begin{tabular}[c]{@{}l@{}}1.High accuracy\\2.Robustness\end{tabular}                                    & \begin{tabular}[c]{@{}l@{}}1.Complexity\\2.Calibration\\3.Synchronization\end{tabular}           \\\hline
\end{tabular}         
\label{table:control_table_1}
\end{table}

\subsection{Object Detection} 
\begin{figure}
\centering  
\includegraphics[width=9cm]{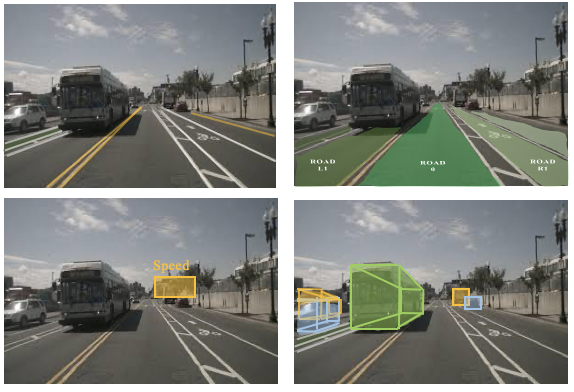}

\caption{The crucial tasks in the object detection for AD. The top-left, top-right, bottom-left, bottom-right figure represents the lane detection, drivable region, traffic sign detection and 3D object detection.}
\label{fig:3D Object Detection}
\end{figure}

The purpose of object detection is to detect the static and dynamic targets in the field of view of the sensors. The results of some detection tasks can be seen in Fig. \ref{fig:3D Object Detection}.

\subsubsection{Lane Detection} 
Lane Detection is to recognize the lane in the views of sensors, to assist driving. For universal process, it involves three sections, including image pre-processing, lane detection, and tracking. The purpose of image pre-processing, such as Region of Interest (RoI) extraction, inverse perspective mapping, and segmentation, is to reduce the computing cost and eliminate noise. The methods of lane detection and tracking can be divided into the Computer Vision based (CV-based) method and the learning-based method \cite{J7_Detection}.

CV-based methods in lane detection are broadly utilized nowadays, primarily because of their light computing cost and easy reproduction. A morphological top-hat transform is utilized to eliminate the irrelevant objects in the field \cite{object_detection_lane_detection_1}. After that, the Hough transform is applied to extract the edge pixel of the image and construct the straight lines. However, the disadvantage is that it is hard to detect the curve lines, so a number of researchers have introduced some effective methods on the Hough transform \cite{object_detection_lane_detection_1}. Some other lines estimation approaches involve the Gaussian Mixture Models (GMM) \cite{object_detection_lane_detection_3}, Random Sample Consensus (RANSAC) \cite{object_detection_lane_detection_4}, Kalman filter \cite{object_detection_lane_detection_5} in complex scenes.

Learning-based methods can be deployed on abundant scenes but they need a great deal of data to train the network with plenty of parameters. \cite{object_detection_lane_detection_8} attempts to design novel multiple sub-headers structures to improve the lane detection performance. To our knowledge, lane detection is integrated into the Advanced Driver Assistance Systems (ADAS) to keep the lane or follow the former vehicle, and researchers pay more attention to 3D lanes \cite{3DLANE}, lanes in closed areas, and unstructured roads.

\subsubsection{Driving Region Detection} 
Driving region detection increases the obstacle information compared to lane detection, which offers the base information for obstacle avoidance function and path planning tasks. We also categorize this task into the CV-based and learning-based approaches.

Driving region detection can be converted to lane detection when the road surface is not obscured by obstacles. Otherwise, it can be seen as a combination of lane detection and 2D target detection. When considering driving region detection as an independent task, it needs to distinguish the road pixel from targets and non-driving regions. The color histogram can meet the requirement and some researchers develop methods on color \cite{object_detection_driving_region_1} and efficiency \cite{object_detection_driving_region_2} to tackle the poor performance on color varying and low effect of it. Region growth methods \cite{ chenlong_tvt} are more robust than the color histogram methods. 

The learning-based methods in driving region detection are similar to image segmentation. For machine learning algorithms, features such as the RGB color, Walsh Hadamard, Histogram of Oriented Gradients (HOG), Local Binary Pattern (LBP), Haar, and LUV channel, can be extracted by the feature extractors and the classification header, such as Support Vector Machine (SVM), Conditional Random Field (CRF), to obtain the final results. The deep neural network can replace the feature extractors and some improvements, such as employing the large visual regions convolutional kernels \cite{object_detection_driving_region_4}, connection by multiple layers \cite{object_detection_driving_region_5}, to achieve competitive performance. We found that learning-based driving region detection results are usually one of the branches of the scene understanding task and researchers attempt to tackle a few challenges including 2D-3D transformation, complex driving regions, etc.

\subsubsection{Traffic Sign Detection}
Traffic signs contain plenty of crucial traffic information, such as road conditions, speed limitations, driving behavior restrictions, and other information. We also divide it into CV-based approaches and learning-based approaches. 

For CV-based detection approaches, the conditions for the approximate color composition of traffic signs in a specific region (in a certain country or city) are similar. In the relatively simple original image, better results can be obtained by threshold separation of specific colors, which can be obtained by adopting the color space distribution, segmentation head, and the SVM classifier \cite{traffic_sign_1}. Some research develops the methods by introducing extra color channels, such as the normal RGB model, the dynamic color threshold, the probability model, and edge information. However, these approaches are hard to solve the problem caused by illumination, fading, occlusion, and bad weather. Some researchers tried to utilize the though transform on the triangle \cite{traffic_sign_8}, circular traffic signs \cite{traffic_sign_9} or a coding gradient detection method \cite{traffic_sign_10}, to handle occlusion and conjunction. The shape-based detection method can solve the problem of in-stable results caused by color change, but it has little capability to overcome the problem caused by occlusion and deformation.

The traffic sign recognition algorithm based on machine learning usually uses the sliding window method to traverse the given traffic sign image. \cite{traffic_sign_12} proposed a variant histogram feature of gradient direction, and trained a single classifier to detect traffic signs through an extreme learning machine. With the continuous research of deep learning algorithms, more and more scholars use deep learning algorithms to detect traffic signs. Readers can regard this classification method as handling feature extraction, including pre-processing and classification \cite{traffic_sign_15}. To extract deeper information, the deeper encoder, integrated Space Pyramid Pooling (SPP) layer, cascaded RCNN network, depth separable convolution, and clipping strategy are introduced to achieve the detection accuracy and high inference speed \cite{traffic_sign_17}. The deep learning method has a satisfactory tolerance for the variation of the color and shape of signs, however, this type of method requires vast amounts of data and manual annotation. Besides, detection networks should have the capability of recognizing different regions with diffident signs and detecting signs over a long distance.

\subsubsection{Visual-based 3D Object Detection}
Visual-based 3D object detection refers to the process of acquiring 3D information (location, dimension, and rotation) about all targets in the field from the image. We divided it into monocular-based and stereo-based detection.

\textit{4.1) Monocular 3D Object Detection:}
Monocular 3D object detection is widely developed and the accuracy has been improving in recent years. Directly associating classification and regression methods, inheriting from 2D object detection networks like \cite{FasterRCNN}, have straightforward structures but perform unsatisfied due to the ill-posed problem of recovering 3D attributions from a single image. There are two main kinds of strategies to solve the shortcoming. 1) Some \cite{MonoPair, Monet3D} introduce the geometric connections between 3D and projected 2D candidates. For example, GS3D \cite{GS3D} decouples the objects into several parts to analyze the surface attributes and instance relationships. Monopair \cite{MonoPair} and Monet3D \cite{Monet3D} consider the relationships between the target and its two nearest neighbors. 2) Besides regressing the 3D bounding candidates, networks also take into account the local or full depth map \cite{CaDDN, DDMP3D} from stereo vision or LiDAR data during the training stage. CaDDN \cite{CaDDN} provides a fully differentiable end-to-end approach for combining depth estimation and object detection tasks. DDMP3D \cite{DDMP3D} utilizes the feature representation of context and depth estimation heads to achieve competitive performance. In addition, \cite{Kinematic3D} introduces successive frames as inputs, which attempts to update the 3D results by associating detection and tracking. Although these methods have no obvious advantages in accuracy, extensive academic research and low cost make them attractive.

\textit{4.2) Stereo 3D Object Detection:} 
Stereo 3D object detection approaches\cite{Stereo-RCNN, OC-Stereo, YOLOStereo3D, Disp-RCNN} are inspired by the parallax analysis from binocular vision. The precise depth value can be reckoned with through the distance between the binocular centers and the associated pair of pixels. Disp-RCNN \cite{Disp-RCNN}, OC-Stereo \cite{OC-Stereo} add segmentation modules paired images from stereo cameras to induce accurate association. YOLOStereo3D \cite{YOLOStereo3D} provides a lightweight model, outperforms a great number of stereo methods based on the complicated disparity convolution operations, and significantly reduces the length of training and testing time. In sum, stereo-based methods could avoid the ill-posed problem of monocular images and are convenient for manufacturers to deploy and maintain in IVs, but accurate measurement on the baseline, the time cost of binocular matching, and the requirement of image preprocessing pose challenges to researchers.

\subsubsection{LiDAR-based 3D Object Detection}
LiDAR-based 3D object detection methods recognize targets 3D properties from point clouds data captured by LiDARs. We categorized it into voxel-wise and point-wise detection.

\textit{5.1) Voxel-wise Object Detection:}
Voxel-wise object detection methods represent the point cloud features in the Birds Eyes View (BEV) and the BEV map is divided into s series of independent voxels manually. The structural design of this type of detection network evolves from point cloud segmentation frameworks, such as PointNet \cite{PointNet} and PointNet++ \cite{PointNet++}, which fit the detection task at the input or output side, and its overall architecture needs to balance performance and efficiency. Taking the classic VoxelNet \cite{voxelnet} and PointPillar \cite{pointpillars} as examples, VoxelNet normalizes the voxels after mapping point clouds, and subsequently applies local feature extraction using several Voxel Feature Encoding (VFE) layers to each non-empty voxel. The voxel-wise features are further abstracted by 3D convolutional middle layers (increasing the receptive field and learning the geometric spatial representation), and finally, the object is detected and classified using a Region Proposal Network (RPN) with position regression.

\textit{5.2) Point-wise Object Detection:}
Point-wise object detection such as \cite{pointrcnn, pvrcnn}, are inspired by PointNet, a classical network for indoor 6D pose estimation with point clouds. Point-RCNN \cite{pointrcnn} is a two-stage point cloud detection framework including candidates generation with semantic segmentation analysis at the first stage and the position revision during the second stage. VoteNet \cite{votenet} extends 2D detection structures to the 3D framework to establish a generic detection framework for point clouds. It basically follows the PointNet++ to reduce the information loss in point cloud transformation. VoteNet also introduces a novel voting mechanism inspired by the Hough transform to locate the targets' centers instead of point on the surface, compared with other 3D networks. It should be noticed that the number of discarded points and modality distinction due to the distance in point clouds detection should be significantly considered for researchers.

\begin{table}[]
\centering
\caption{The Performance of 3D object detection methods in KITTI}
\begin{tabular}{lcccc}
\hline
Method                         & Sensors & Moderate(\%)                        & Easy(\%)                            & Hard(\%)                            \\ \hline
VPFNet\cite{VPFNet}                         & Cam+LiD &  83.21 & 91.02 & 78.20 \\
DVF\cite{DVF}]                            & Cam+LiD & 82.45 &  89.40 & 77.56 \\
CLOCs\cite{CLOCs}                          & Cam+LiD & 82.28 &  89.16 &  77.23 \\
F-ConvNet\cite{F-ConvNet}                      & Cam+LiD & 76.39 &  87.36 &  66.69 \\
Point-RCNN\cite{pointrcnn}                     & LiDAR   &  75.64 &  86.96 &  70.70 \\
PointPillars\cite{pointpillars}                   & LiDAR   &  74.31 &  82.58 &  68.99 \\
F-PointNet\cite{F-PointNet}                     & Cam+LiD &  69.79 &  82.19 &  60.59 \\
AVOD\cite{AVOD}                           & Cam+LiD &  66.47 &  76.39 &  60.23 \\
MV3D\cite{MV3D}                           & Cam+LiD &  63.63 &  74.97 &  54.00 \\
Disp-RCNN\cite{Disp-RCNN}                     & Stereo  & 43.27 &  67.02 &  36.43 \\
YOLOStereo3D\cite{YOLOStereo3D}                   & Stereo  &  41.25 &  65.68 &30.42 \\
OC-Stereo\cite{OC-Stereo}                      & Stereo  &  37.60 &  55.15 &  30.25 \\
Stereo-RCNN\cite{Stereo-RCNN}                   & Stereo  &  30.23 &  47.58 &  23.72 \\
MonoDETR\cite{MonoDETR}                       & Mono    &  16.26 &  24.52 &  13.93 \\
MonoDTR\cite{MonoDTR}                        & Mono    &  15.39 &  21.99 &  12.73 \\
CaDDN\cite{CaDDN}                          & Mono    &  13.41 &  19.17 & 11.46 \\
DDMP-3D\cite{DDMP3D} & Mono    &  12.78 &  19.71 &  9.80  \\
GS3D\cite{GS3D}                           & Mono    &  2.90 &  4.47  &  2.47 \\ \hline
\end{tabular}
\label{table:experiment_1}
\end{table}

\subsubsection{Fusion-based 3D Object Detection}
LiDARs, radars, and cameras are widely deployed in IVs for perception tasks and combination of these types of sensors could make the vehicles robust and able to detect targets full-time. However, this does not mean  that fusion-based methods will outperform the approaches with a single sensor. There are two main reasons for the disadvantage of fusion-based methods. 1) It is challenging for the network to fill the modalities gap from various sensors; 2) The system error and measurement errors such as from calibration and synchronization are hard to eliminate and they would be propagated and amplified in the networks. Most researchers propose solutions to handle these difficulties and achieve some competitive outcomes. In this section, we categorize the fusion-based objection detection task based on the types of sensors.

\textit{6.1) Camera and LiDAR:} 
Cameras and LiDARs are two crucial sensors for AD and researcher firstly focus on fusion parallel methods, which extract point clouds and images information at the same time. MV3D \cite{MV3D} and AVOD \cite{AVOD} utilize the shared 3D anchors on the point cloud and the corresponding images. ContFuse \cite{ContFuse} and MMF \cite{MMF} adopt tightly-coupled fusion approaches with a consecutive fusion layer. 3D-CVF \cite{3D-CVF} introduces a cross-view spatial feature fusion method to fuse the images and point clouds. In addition, EPNet \cite{EPNet} focuses on the point cloud system and projects the images on it with point-based strategy on the geometric space.

Compared with parallel approaches, sequential methods are readable and deployable because of no need to introduce association structures to reduce the gaps. F-PointNet \cite{F-PointNet} and F-ConvNet \cite{F-ConvNet} attempt to reduce the searching areas by generating 3D bounding boxes within 2D candidates. PointPainting \cite{PointPainting} outputs semantic information and projects each point on the corresponding point to improve 3D object detection accuracy. CLOCs \cite{CLOCs} fuses the data after the independent extractors and achieve a competitive result on KITTI \cite{dataset_KITTI}. DVF \cite{DVF} adopts the 2D truth as guidance and then extract 3D properties by the point clouds.

\textit{6.2) Camera and Radar:} 
Combining the images and data from Radars can effectively reduce the cost and maintain accuracy. \cite{object_detection_fusion_3D_cameraradar_1} projects radar detection results to the image space and utilize them to boost the object detection accuracy for distant objects. CRF-Net \cite{object_detection_fusion_3D_cameraradar_3} develops the method with a vertical presentation.

\textit{6.3) LiDAR and Radar:} 
This type of fusion focuses on extremely harsh weather conditions and distinct targets. RadarNet \cite{RadarNet} fuses radar and LiDAR data via a novel-based early fusion approach. It leverages the radar's long sensing range via an attention-based fusion. MVDNet \cite{MVDNet} generates proposals from two sensors and then fuses region-wise features between multi-modal sensor streams to improve final detection results. ST-MVDNet \cite{ST-MVDNet} develops the structure by enforcing output consistency between a Teacher network and a Student network, and introducing missing modalities to tackle the degeneration problem when one type of data is missing.

\textit{6.4) Camera, LiDAR and Radar:} 
In this fusion type, researchers attempt to design a robust perception system in different weather conditions. \cite{lidar_camera_radar} obtains object detection outputs with a PointNet \cite{PointNet} architecture by projecting the images onto the point cloud directly. Parallel to the previous frame, the point cloud from the radar is processed to predict velocity which is then associated with the final detection output. RVF-Net \cite{RVF-Net} fuses all of the data on the input procedure and achieves satisfying results on the nuScenes \cite{nuScenes} data set. 

\textit{6.5) Others:} 
Ultrasonic radar judges the distance of obstacles through the time of sound transmission in the air, and the accuracy can achieve a centimeter scale within 5 meters. This sensor is mostly used in autonomous parking scenes. An infrared camera with an infrared lamp can capture the infrared spectral characteristics to achieve the effect of night vision imaging. Besides, research on event cameras is one of the hot topics nowadays. Event cameras process data based on pipeline timestamps, rather than processing individual pixels in a frame plane. Because the data has the nature of time sequence, the traditional network structure can not process the data, so how to fuse with other sensors will be one of the research points in the future.

The performance of 3D object detection methods with various combination of different sensor types in KITTI \cite{dataset_KITTI} is shown in TABLE \ref{table:experiment_1}. Here, KITTI divides whole data into three evaluation scenes (easy, moderate and hard) through the frame's complexity and computes 3D-AP, an extended method from 2D-AP \cite{dataset_seg_VOC} on these three scenes. We summarize: 1) Adopting fusion strategies could achieve competing results for 3D object detection tasks mainly because of introducing more initial information. But this type of method requires researchers to eliminate or reduce modal differences. 2) Due to the characteristics of sensors, limited resolutions of cameras, and the definition of the reference system in KITTI, the performance of visual-based methods is weaker than LiDAR-based. However, visual-based methods attract a number of researchers because of their maintainability, economy, and easy deployment. 3) The self-attention mechanism (Transformer structure) and BEV method \cite{VPFNet, MonoDETR, MonoDTR} could improve the accuracy of cross-modality fusion, feature extraction, etc. In addition, to address data hungry and model robustness, current research studies train and test models on additional data such as unScenes\cite{nuScenes}, Waymo\cite{Waymo}.

\subsection{Scene Understanding}
We define scene understanding in our paper as the multiple outputs for each pixel or point instead of each target. This section, we divide it into three sub-sections, segmentation, depth, and flow estimation. We only focus on academic research and applications in AD areas.

\subsubsection{Segmentation in Autonomous Driving}
The target of semantic segmentation is to partition a scene into several meaningful parts, usually by labeling each pixel in the image with semantics (semantic segmentation), by simultaneously detecting objects and distinguishing each pixel from each object (instance segmentation), or by combining semantic and instance segmentation (panoptic segmentation) \cite{survey_dynamic_20}. The segmentation is one of the crucial tasks in computer vision and researchers evaluate their models on ADE20K \cite{dataset_seg_ADE20K}, Pascal-VOC \cite{dataset_seg_VOC}, CityScape \cite{dataset_seg_CityScape}, etc. However, in AD scenarios, the classic 3D CV area, it is hard to complete the perception task independently. It is usually involved in lane detection, driving region detection, visual interface module, or combined with point clouds to provide semantic information. We will briefly introduce the general background based on segmentation, and then highlight segmentation research on AD.

\textit{1.1) Semantic Segmentation:} 
Fully Convolutional Network (FCN) \cite{scene_FCN} is a popular structure for semantic segmentation which adopts convolutional layers to recover the size of output maps. Some work extends FCN by introducing an improved encoder-decoder \cite{scene_segnet}, the dilated convolutions \cite{scene_dilated}, CRFs \cite{scene_DeepLabv1}, atrous spatial pyramid pooling(ASPP) \cite{scene_DeepLabv2}. In addition, the above approaches attend to fixed, square context regions because of the pooling and dilation convolution operations. The relational context method \cite{scene_OCRNet} extracts the relationship between pixels. \cite{scene_UNet} pursues high resolution by channel concatenation and skip connection, especially in the medical field. In the field of AD, the semantic segmentation networks may be familiar with the common structures, and researchers should pay more attention to the special categories, and occlusion, and evaluate their models on data sets of road scenarios \cite{dataset_seg_CityScape}. To achieve SOTA results on data sets, the researcher introduces the multiple scale attention mechanism \cite{scene_multi_scale_attention}, boundary-aware segmentation module \cite{scene_boundary_aware_segmentation}. Besides, some research focuses on the targets' attributes on roads like considering the intrinsic relevance among the cross-class objects \cite{scene_intrinsic_relevance} or semi-supervised segmentation mechanism because of the lack of labeled data on AD scenarios.

\begin{table}[]
\centering
\caption{Results on CityScape for panoptic segmentation and KITTI for depth estimation}
\begin{tabular}{lccc}
\hline
Method        & PQ(\%)                                                  & SQ(\%)                                                  & RQ(\%)                                                  \\ \hline
Axial-D\cite{scene_Axial_deeplab}] & 62.7 & 82.2 & 75.3 \\
TASC\cite{scene_TASC}          &  60.7 &  81.0 &  73.8 \\ \hline
Method        & SILog(log(m))                                              & sqErrorRel(\%)                                          & iRMSE(1/km)                                               \\ \hline
BANet\cite{scene_BANet}         & 0.1155                                               & 2.31                                                & 12.17                                               \\
VNL\cite{scene_From_big_to_small}           & 0.1265                        & 2.46                         & 13.02                                               \\
SDNet\cite{scene_SDNet}]         & 0.1468                        & 3.90                         & 15.96                                               \\
MultiDepth\cite{scene_Multidepth}    & 0.1605                        &  3.89                        & 18.21                                               \\ \hline
\end{tabular}
\label{table:experiment_2}
\end{table}

\textit{1.2) Instance Segmentation:} 
Instance segmentation is to predict a mask and its corresponding category for each object instance. Early method \cite{scene_Mask_RCNN} designs an architecture to realize both object detection and segmentation missions. Mask-RCNN \cite{scene_Mask_RCNN} extends Faster-RCNN to identify each pixel's category with binary segmentation and pools image features from Region of Interest (RoI) following a Region Proposal Network (RPN). Some researchers develop the base structure by introducing a coefficients network \cite{scene_Explicit}, the IoU score for each mask, and shape priors to refine predictions. Similar with the 2D object detection methods, \cite{scene_Blendmask} replaces the detectors with the one-stage structures. \cite{scene_Solov2} attempts to avoid the effect of detection and achieve remarkable performances. To achieve competitive segmentation results on AD datasets, researchers focus on the geometric information on 3D space \cite{scene_Polytransform}, boundary recognition \cite{scene_Look_closer_to_segment_better}, combining the semantic segmentation (panoptic segmentation) \cite{scene_Scaling_wide} or intruding multiple frames (video-base) \cite{scene_Naive_student}.

\textit{1.3) Panoptic Segmentation:} Panoptic segmentation is proposed to unify pixel-level and instance-level semantic segmentation \cite{scene_Panoptic_segmentation}, and \cite{scene_Panoptic_FPN} designs a different branch to regress the semantic and instance segmentation results. Panoptic-FCN \cite{scene_Panoptic_FCN} aims to represent and predict foreground things and background stuff in a unified fully convolutional pipeline. Panoptic SegFormer \cite{scene_Panoptic_SegFormer} introduces a concise and effective framework for panoptic segmentation with transformers. For AD scenarios, TASC \cite{scene_TASC} proposes a new differentiable approach to reduce the gap between the two sub-tasks during training. Axial-DeepLab \cite{scene_Axial_deeplab} builds a stand-alone attention model with a global receptive field and a position-sensitive attention layer to capture the positional information with low computational cost. Besides, researchers address the multiple scales on roads by introducing a novel crop-aware bounding box regression loss and a sample approach \cite{scene_Improving_panoptic_segmentation}, and capture the targets' boundary by a combinatorial optimization strategy. These methods achieve competitive results on the task of CityScape \cite{dataset_seg_CityScape} or Mapillary Vistas \cite{dataset_seg_Mapillary_Vistas}.
\subsubsection{Depth Estimation in Autonomous Driving}
This type of task is to present the depth information on the camera plane, which is an effective way to enhance the visual-based 3D object detection and a potential bridge to connect the LiDAR and camera.

The depth completion task is a sub-problem of depth estimation \cite{scene_Depth_completion}. In the sparse-to-dense depth completion problem, researchers infer the dense depth map of a 3D scene from a sparse depth map by computational methods or multiple data from sensors. The main difficulties include: 1) the irregularly spaced pattern in the sparse depth, 2) the fusion methods for multiple sensor modalities (optional), and 3) the lack of dense pixel-level ground truth for some data and the real world (optional).

Depth estimation is the task of measuring the distance of each pixel relative to the camera. The depth value is extracted from either monocular or stereo images with supervised (the dense map obtained by depth completion) \cite{scene_Indoor_scene_structure_analysis}, unsupervised \cite{scene_Semi_supervised_deep_learning}, LiDAR guidance \cite{scene_Refinedmpl} or stereo computing \cite{scene_Deepstereo}. Some approaches \cite{scene_Multi_scale_continuous_crfs, scene_Pad_Net} introduce the CRF module, multi-tasks structure, global extractor, and the piece-wise planarity priors to achieve competitive performances in popular benchmarks such as KITTI \cite{dataset_KITTI} and NYUv2 \cite{dataset_NYUv2}. Models are typically evaluated according to an RMS metric \cite{dataset_KITTI}.

For outdoor monocular depth estimation, DORN \cite{scene_DORN} adopts a multi-scale network structure to capture the contextual information. MultiDepth \cite{scene_Multidepth} makes use of depth interval classification as an auxiliary task. HGR \cite{scene_HGR} proposes a hierarchical guidance and regularization learning framework to estimate the depth. SDNet \cite{scene_SDNet} improves the results by utilizing a dual independent estimation head involving depth and semantics. VNL \cite{scene_From_big_to_small} designs a novel structure that includes local planar guidance layers at multiple stages. \cite{scene_Enforcing_geometric_constraints} uses the geometric constraints of normal directions determined by randomly sampled three points to improve the depth prediction accuracy. BANet \cite{scene_BANet} introduces bidirectional attention modules which adopt the feed-forward feature maps and incorporate the global information to eliminate ambiguity. The Unsupervised method \cite{scene_monodepth2} attracts plenty of researchers because it could reduce the requirements on the labeled data and eliminate the over-fit problem. In addition, the pure monocular depth estimation only obtains the relative depth value because of the ill-posed problem, and the stereo guidance methods could obtain the absolute depth value. \cite{ scene_DepthFormer} introduces the Transformer structures to achieve competitive results. The stereo depth estimation methods can be found in the stereo disparity estimation task.

\subsubsection{Flow Estimation in Autonomous Driving} 
Similar to the segmentation and depth estimation tasks, flow estimation focuses on the image plane and it presents the pixel movement during a data frame. It attracts interest nowadays and its research can be used in event camera methods.

\textit{3.1) Optical Flow Estimation:}
Optical flow refers to the pixels' movement in the imaging system including two directions, the horizontal and vertical. Similar to unsupervised video-based depth estimation, the pixel motion \cite{scene_flownet} can be deduced by minimizing differences between the target and source images. SPyNet \cite{scene_spynet} proposes a lightweight framework that adopts classical spatial-pyramid formulation for optical flow estimation. In addition, it attempts to estimate large-displacement movement and accurate sub-pixel flow. PWC-Net \cite{scene_pwcnet} includes three sub-nets, the feature pyramid extractor, warping layer, and cost volume layer, to improve the quality of optical flow. 

\textit{3.2) Scene Flow Estimation:} 
Scene flow estimation indicates a 3-dimensional movement field which can be treated as the extension of optical flow. Therefore, it is the combination of optical flow and depth estimation in 3D scenarios. Monocular images are seldom utilized in the holistic training step for scene flow, and the structure takes the binocular videos as input to regress disparity to restore the scale. DRISF \cite{scene_drisf} treats the inference step of Gaussian Newton (GN) as a Recurrent Neural Network (RNN) which means it can be trained in an end-to-end method. FD-Net \cite{scene_dfnet} further extends the unsupervised depth estimation and disentangles the full flow into object flow (targets pixels) and rigid flow (background pixels) to assess the characteristics respectively, which is able to avoid the warping ambiguity due to the occlusion and truncation. Competitive Collaboration (CC) \cite{scene_cc} sets the scene flow estimation as a game with three players. Two of them compete for the resource and the last one acts as a moderator. GeoNet \cite{scene_geonet} consists of two modules, a monocular depth with the 6 DoF ego-motion estimation, and a residual network to learn the object's optical flow.

The performance of panoptic segmentation and depth estimation on CityScape and KITTI is shown in TABLE \ref{table:experiment_2}. PQ, SQ, RQ refer the panoptic segmentation, segmentation quality, and recognition quality respectively in \cite{scene_Panoptic_segmentation}, and  for depth estimation, SILog (Scale invariant logarithmic error), sqErrorRel (Relative squared error), and iRMSE (Root mean squared error of the inverse depth) are classical metrics in KITTI. Similar to detection, researchers introduce the self-attention mechanism, extra training data and novel network units to develop the accuracy in scene understanding tasks. And we mention that above tasks do not directly provide outputs to the downstream tasks such as planning and control in AD. In the actual tasks, semantic segmentation, depth estimation and optical flow estimation will be combined with each other to provide richer pixel semantic information, so as to improve the accuracy of cross-modality data fusion, spatial detection and tracking for moving targets.




\subsection{Prediction}
In order to safely and efficiently navigate in complex traffic scenarios, an AD framework should be able to predict the way in which the other traffic agents (such as vehicles and pedestrians) will behave in the near future.
Prediction can be defined as probable results according to past perceptions. Let $X_{t}^{i}$ be a vector with the spatial coordinates of agents  $i$ at observation time $t$, with $t\in \left \{ X_{1}^{i},X_{2}^{i}, ... .X_{T_{obs}}^{i}  \right \} $.

\subsubsection{Model-based Approaches} 
These methods predict the behaviors of agents, such as changing lanes, turning left, and so on. One of the simplest methods to predict the probability distribution of vehicle behavior is the autonomous multiple models (AMM) algorithm. This algorithm computes the maximum probability trajectory of each agent.

\subsubsection{Data-driven Approaches}
These methods are mainly composed of the neural network. After training on the perception dataset, the model makes a prediction of the next behavior. DESIRE \cite{prediction1} proposes an encoder-decoder framework that innovatively incorporates the scenario context and the interactions between traffic agents.  SIMP \cite{prediction2} discretizes the output space, calculates the distribution of the vehicle's destination, and predicts an estimated time of arrival and a spatial offset. FaF \cite{prediction3} pioneers the unification of detection and short-term motion forecasting based on LiDAR point clouds. The prediction module is sometimes separated from the perception, mainly because the downstream planning module receives both the perception and the prediction results. Future research on prediction will focus on the formulation of generalized rules, the universality of scenarios and the simplicity of modules.

\subsection{Tracking}

The tracking problem begins with a sequence of vehicle-mounted sensor data. Depending on if neural network is embedded in the tracking framework, we divide them into the traditional method and the neural network method.

\subsubsection{Tradition Method}
The Kalman filter\cite{Kalman1} is a famous algorithm, particularly with regard to tracking agents. Because of the low computational cost, the Kalman-based method \cite{Kalman2} has quick response time even on low-spec hardware in simple scenarios.

The tracking problem also can be shown as a graph search problem \cite{graph1}.  Compared with Kalman-based methods, The most important advantage of graph-based approach is that it is better for the multi-tracking problems. \cite{graph2} exploits graph-based methods using the min-cost approach to solve tracking problems.

\subsubsection{Neural Network Method}
Neural networks have the advantage of being able to learn important and robust features given training data that is relevant and with sufficient quantity.

CNN is widely used in agents tracking. \cite{TrackingNetwork1} handles multi-agent tracking using combinations of values from convolutional layers. \cite{TrackingNetwork2} proposes appropriate filters for information drawn from shallow convolutional layers, achieving the same level of robustness compared with deeper layers or a combination of multiple layers.

RNN also provides a smart method to solve temporal coherence problem in tracking task. \cite{TrackingLSTM1} uses an LSTM-based classifier to track agents across multiple frames in time. Compared with CNN method, the LSTM-based approach is better suited to remove and reinsert candidate observations particularly when objects leave or reenter the visible area of the scene. Joint perception and tracking can achieve the SOTA results in these two tasks. In reality, stable tracking can reduce the requirements of the system for real-time detection and can also correct the detection results. At present, the strategy of joint task learning has been favored by more and more researchers.


\section{Planning}

The planning module is responsible for finding a local trajectory for the low-level controller of the ego vehicle to track. 

\begin{figure}
\centering  
\includegraphics[width=8.5cm]{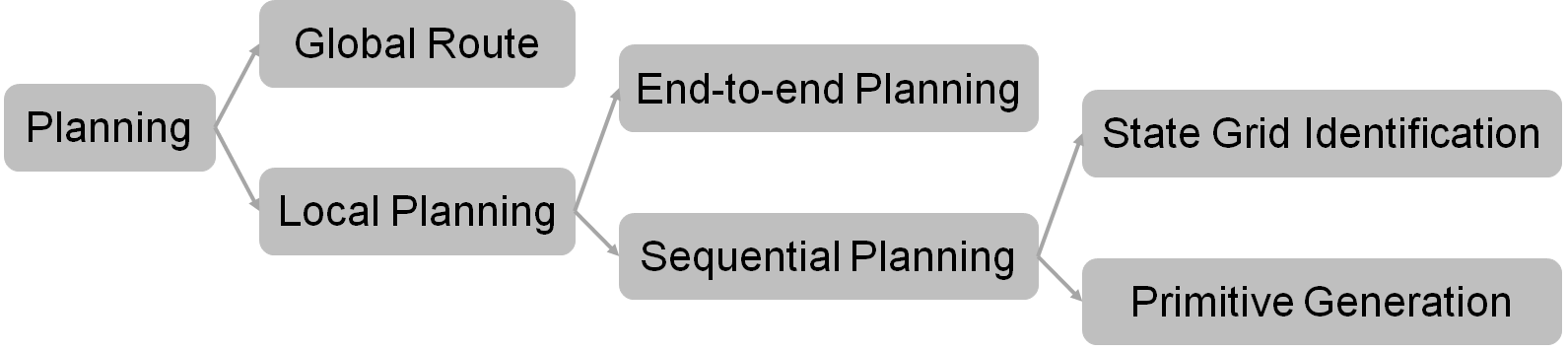}

\caption{The structure of planning methodology}
\label{fig:planning methodology}
\end{figure}

The planning module is responsible for finding a local trajectory for the low-level controller of the ego vehicle to track. Herein, “local” means that the resultant trajectory is short in its spatial or temporal range; otherwise the ego vehicle cannot react to risks beyond the sensor ranges. The planning module typically contains three functions, namely global route planning, local behavior planning, and local trajectory planning \cite{survey_planning_1}. Global route planning provides a road-level path from the start point to the destination on a global map; local behavior planning decides a driving action type (e.g., car-following, nudge, side pass, yield, and overtake) for the next several seconds while local trajectory planning generates a short-term trajectory based on the decided behavior type. This section reviews the techniques related to the three functions in the planning module as Fig. \ref{fig:planning methodology}.

\subsection{Global Route Planning}
Global route planning is responsible for finding the best road-level path in a road network, which is presented as a directed graph containing millions of edges and nodes. A route planner searches in the directed graph to find the minimal-cost sequence that links the starting and destination nodes. Herein, the cost is defined based on the query time, preprocessing complexity, memory occupancy, and/or solution robustness considered.

The development history of global route planning techniques is much longer than that of autonomous vehicle techniques because global route planning also serves manually driven vehicles. As indicated by \cite{planning_4}, the existing global route planning methods are classified as goal-directed methods, separator-based methods, hierarchical methods, bounded-hop methods, and their combinations.

\subsection{Local Behavior/Trajectory Planning}

\begin{figure}
\centering  
\includegraphics[width=8.5cm]{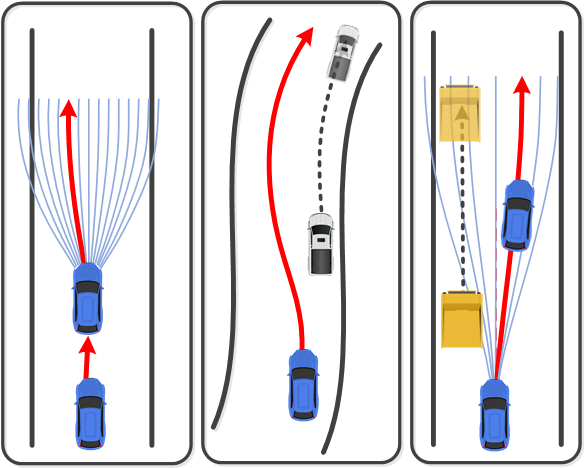}

\caption{The route section for local behavior planning}
\label{fig:local planning}
\end{figure}

Local behavior planning and local trajectory planning functions work together to output a local trajectory along the identified global route as shown in Fig. \ref{fig:local planning}. Since the resultant trajectory is local, the two functions have to be implemented in a receding-horizon way unless the global destination is not far away \cite{planning_6}. It deserves to emphasize that the output of the two functions should be a trajectory rather than a path \cite{planning_7}, otherwise extra efforts are needed for the ego vehicle to evade the moving obstacles in the environment.

Broadly speaking, the two functions would work in two different ways. One is the end-to-end way, i.e., to develop an integrated system that receives the raw data from the on-board sensors and outputs a local trajectory directly. The other way is to implement the local behavior planning and local trajectory planning functions sequentially.

\subsubsection{End-to-end Methods} 
Compared with the sequential-planning solution reviewed in the next subsection, an end-to-end solution nominally deals with vehicle-environment interactions more efficiently because there is not an external gap between the perception and planning modules\cite{Motionplanning}. The input of an end-to-end system is the large amount of raw data obtained by the on-board sensors whereas the output is a local trajectory. Since the relationship between the input and output is too intricate to be summarized as complete rules \cite{planning_9}, machine learning methods are commonly used, most of which are classified as imitation-learning-based and reinforcement-learning-based methods \cite{planning_10}.

An imitation-learning-based method builds a neuro network based on training samples \cite{planning_11,Hierarchical}. Challenges lie in how to collect massive training samples that are consistent and how to guarantee learning efficiency (e.g., free from overfitting). Reinforcement-learning-based methods obtain knowledge by trial-and-error operations, thus they rely less on the quality and quantity of external training samples \cite{planning_13}.

End-to-end methods are still not mature, thus most of them are trained/tested in simulations rather than real-world scenarios . Recent research efforts focus on how to enhance learning interpretability, security, and efficiency.

\subsubsection{Sequential-planning-based Methods} 
As opposed to the aforementioned end-to-end solution, applying local behavior planning and trajectory planning functions sequentially has been a common and conventional choice in the past decade. However, the boundary between local behavior planning and trajectory planning is rather blurred \cite{survey_planning_1}, e.g., some behavior planners do more than just identify the behavior type. For the convenience of understanding, this paper does not distinguish between the two functions strictly and the related methods are simply regarded as trajectory planning methods.

Nominally, trajectory planning is done by solving an optimal control problem (OCP), which minimizes a predefined cost function with multiple types of hard constraints satisfied \cite{planning_16}. The solution to the OCP is presented as time-continuous control and state profiles, wherein the desired trajectory is reflected by (part of) the state profiles. Since the analytical solution to such an OCP is generally not available, two types of operations are needed to construct a trajectory. 

Concretely, the first type of operation is to identify a sequence of state grids while the second type is to generate primitives between adjacent state grids.

\textit{2.1) State Grid Identification:}
State grid identification can be done by search, selection, optimization, or potential minimization. Search-based methods abstract the continuous state space related to the aforementioned OCP into a graph and find a link of states there. Prevalent search-based methods include A* search \cite{ planning_18} and dynamic programming (DP) \cite{planning_19}. Selection-based methods decide the state grids in the next one or several steps by seeking the candidate with the optimal cost/reward function value. Greedy selection \cite{planning_22} and Markov decision process (MDP) series methods typically \cite{planning_23} fall into this category. An optimization-based method discretizes the original OCP into a mathematical program (MP), the solution of which are high-resolution state grids \cite{planning_25, planning_29}. MP solvers are further classified as gradient-based and non-gradient-based ones;  gradient-based solvers typically solve nonlinear programs \cite{planning_16, planning_29}, quadratic programs \cite{planning_21, planning_31}, quadratically constrained quadratic programs \cite{planning_30} or mix-integer programs \cite{planning_32}; non-gradient-based solvers are typically represented by metaheuristics \cite{planning_33}. Potential-minimization-based methods adjust the state grid positions by simulating the process they are repelled or attracted by forces or in a heuristic potential field. Prevalent methods in this category include the elastic band (EB) series \cite{planning_35}, artificial potential field methods \cite{planning_7}, and force-balance model \cite{planning_37}. 

The capability of each state grid identification method is different. For example, gradient-optimization-based and potential-minimization-based methods are generally more flexible and stable than typical search-/selection-based methods \cite{planning_38}, but search-/selection-based methods are more efficient to explore the entire state space globally \cite{planning_35, planning_39, planning_40}. Different methods could be combined jointly as a coarse-to-fine strategy, as has been implemented by many studies such as \cite{planning_16, planning_21, planning_29, planning_30}.

\textit{2.2 Primitive Generation:}
Primitive generationis commonly done by closed-form rules, simulation, interpolation, and optimization. Closed-form rules refer to methods that generate primitives by analytical methods with closed-form solutions. Typical methods include the Dubins/Reeds-Shepp curves \cite{planning_42}, polynomials \cite{planning_22}, and theoretical optimal control methods \cite{planning_44}. Simulation-based methods generate trajectory/path primitives by forward simulation, which runs fast because it has no degree of freedom \cite{planning_45}. Interpolation-based methods are represented by splines or parameterized polynomials. Optimization-based methods solve a small-scale OCP numerically to connect two state grids \cite{planning_47}.

State grid identification and primitive generation are two necessary operations to construct a trajectory. Both operations may be organized in various ways. For example, \cite{planning_45} integrates both operations in an iterative loop; \cite{planning_47} builds a graph of primitives offline before online state grid identification; \cite{planning_21} identifies the state grids before generating connective primitives.

If a planner only finds a path rather than a trajectory, then a time course should be attached to the planned path as a post-processing step \cite{planning_49}. This strategy, denoted as path velocity decomposition (PVD), has been commonly used because it converts a 3D problem into two 2D ones, which largely facilitates the solution process. Conversely, non-PVD methods directly plan trajectories, which has the underlying merit to improve the solution optimality \cite{planning_19, planning_20, planning_50}.

Recent studies in this research domain include how to develop specific planners that fit specific scenarios/tasks particularly \cite{planning_6}, and how to plan safe trajectories with imperfect upstream/downstream modules \cite{planning_55}.








\section{conclusion}
This article is the third part of our work (Part \uppercase\expandafter{\romannumeral2} for the technology survey). In this paper, we provide a review of wide introductions on research development with milestones of perception and planning in AD and IVs. In addition, we provide a few experiment results and unique opinions for these two tasks. In combination with the other two parts, we expect that our whole work will bring novel and diverse insights to researchers and abecedarians, and serve as a bridge between past and future.

\small\bibliographystyle{IEEEtran}
\bibliography{body/mylib.bib}


\section{Biography Section}

\begin{IEEEbiography}[{\includegraphics[width=1in,height=1.25in,clip,keepaspectratio]{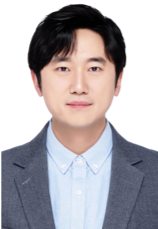}}]{Long Chen} (Senior Member, IEEE) received the Ph.D. degree in electrical and electronic engineering from Wuhan University in 2013.

He is currently a Professor with State Key Laboratory of Management and Control for Complex Systems, Institute of Automation, Chinese Academy of Sciences, Beijing, China. He is also with the presidents office, Waytous Ltd., Beijing. His research interests include autonomous driving, robotics, and artificial intelligence, where he has contributed more than 100 publications.

Prof. Chen serves as an Associate Editor for the IEEE TRANSACTIONS ON INTELLIGENT TRANSPORTATION SYSTEMS, the IEEE/CAA JOURNAL OF AUTOMATIC SINICA, the IEEE TRANSACTIONS ON INTELLIGENT VEHICLES and the IEEE Technical Committee on Cyber-Physical Systems.
\end{IEEEbiography}
\begin{IEEEbiography}[{\includegraphics[width=1in,height=1.25in,clip,keepaspectratio]{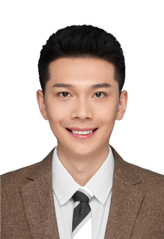}}]{Siyu Teng} received the M.S. degree in computer science and engineering from Jilin University, Changchun, China, in 2021. He is currently pursuing the Ph.D. degree with the Department of Computer Science, Beijing Normal University-Hong Kong Baptist University United International College, Zhuhai, China, and also with the Department of Computer Science, Hong Kong Baptish University, Hong Kong.

His main interests are end-to-end autonomous driving and interpretable deep learning.
\end{IEEEbiography}


\begin{IEEEbiography}[{\includegraphics[width=1in,height=1.25in,clip,keepaspectratio]{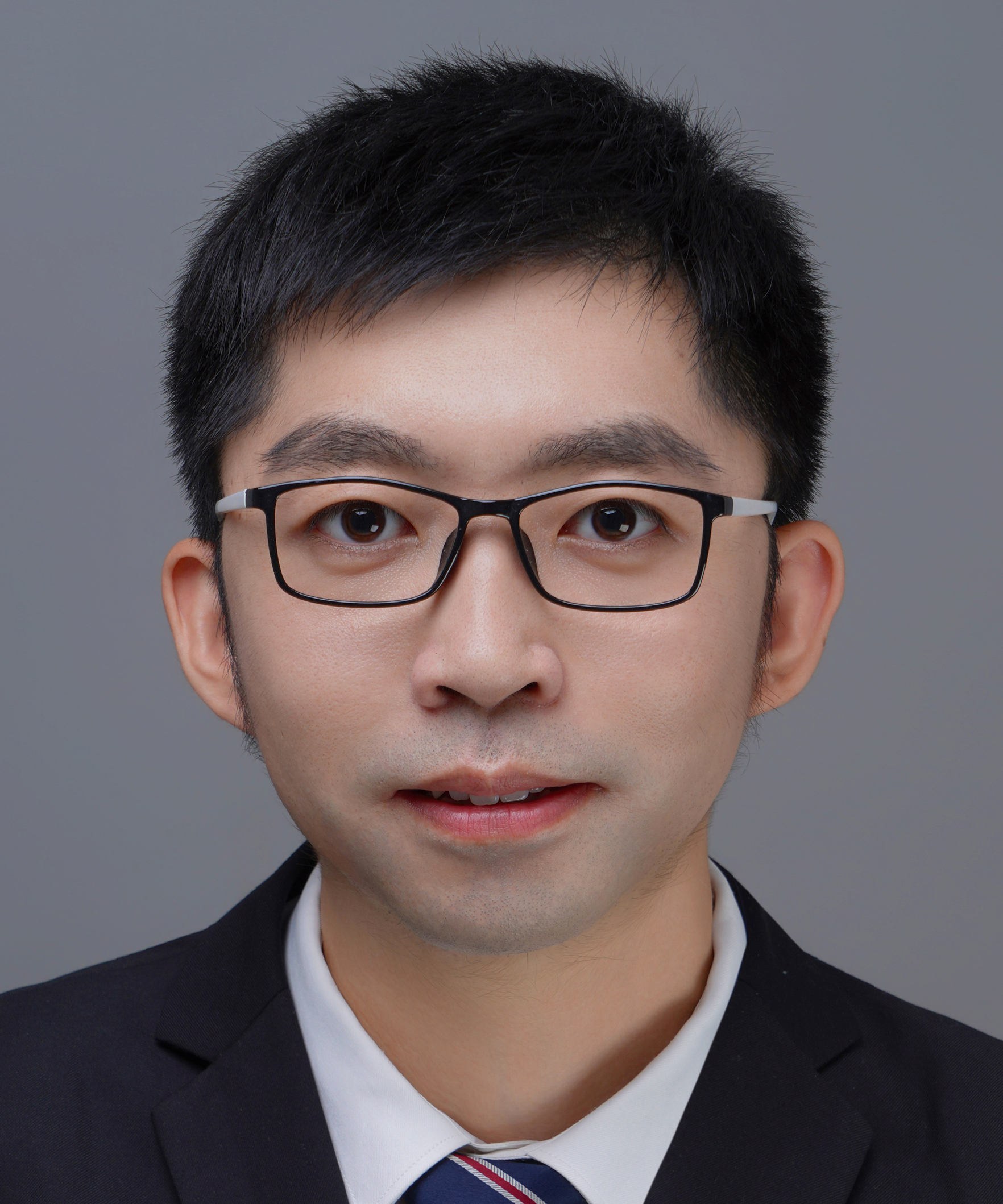}}]{Bai Li} received his B.S. degree in 2013 from Beihang University, China, and the Ph.D. degree in 2018 from the College of Control Science and Engineering, Zhejiang University, China. 
He is currently an associate professor in the College of Mechanical and Vehicle Engineering, Hunan University, Changsha, China. Before joining Hunan University, he was a research engineer of JD.com Inc., Beijing, China from 2018 to 2020. Prof. Li was the first author of more than 70 journal/conference papers and two books in numerical optimization, optimal control, and trajectory planning. 

Dr. Li was a recipient of the International Federation of Automatic Control (IFAC) 2014–2016 Best Journal Paper Prize. He is an Associate Editor of IEEE TRANSACTIONS ON INTELLIGENT VEHICLES.
\end{IEEEbiography}

\begin{IEEEbiography}[{\includegraphics[width=1in,height=1.25in,clip,keepaspectratio]{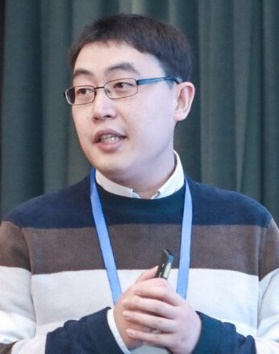}}]{Xiaoxiang Na}received the B.Sc. and M.Sc. degrees in automotive engineering from the College of Automotive Engineering, Jilin University, China, in 2007 and 2009, respectively. He received the Ph.D. degree in driver-vehicle dynamics from the Department of Engineering, University of Cambridge, U.K. in 2014.

He is currently a University Assistant Professor in Applied Mechanics at the Department of Engineering, University of Cambridge. From 2014 to 2023, he worked as a Research Associate and later Senior Research Associate at the Centre for Sustainable Road Freight (SRF) at the University of Cambridge. His main research interests include driver-vehicle dynamics, in-service monitoring of heavy goods vehicle (HGV) operations, and evaluation and modelling of energy performance of HGVs.
\end{IEEEbiography}

\begin{IEEEbiography}[{\includegraphics[width=1in,height=1.25in,clip,keepaspectratio]{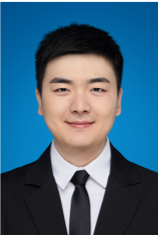}}]{Yuchen Li} received the B.E. degree in software engineering from the University of Science and Technology Beijing, Beijing, China, in 2016, and the M.E. degree in software engineering from Beihang University, Beijing, in 2020. He is currently pursuing the Ph.D. degree with the Department of Computer Science, Beijing Normal University-Hong Kong Baptist University United International College, Zhuhai, China, and also with the Department of Computer Science, Hong Kong Baptish University, Hong Kong.

His research interests cover computer vision, 3-D object detection, and
autonomous driving.
\end{IEEEbiography}

\begin{IEEEbiography}[{\includegraphics[width=1in,height=1.25in,clip,keepaspectratio]{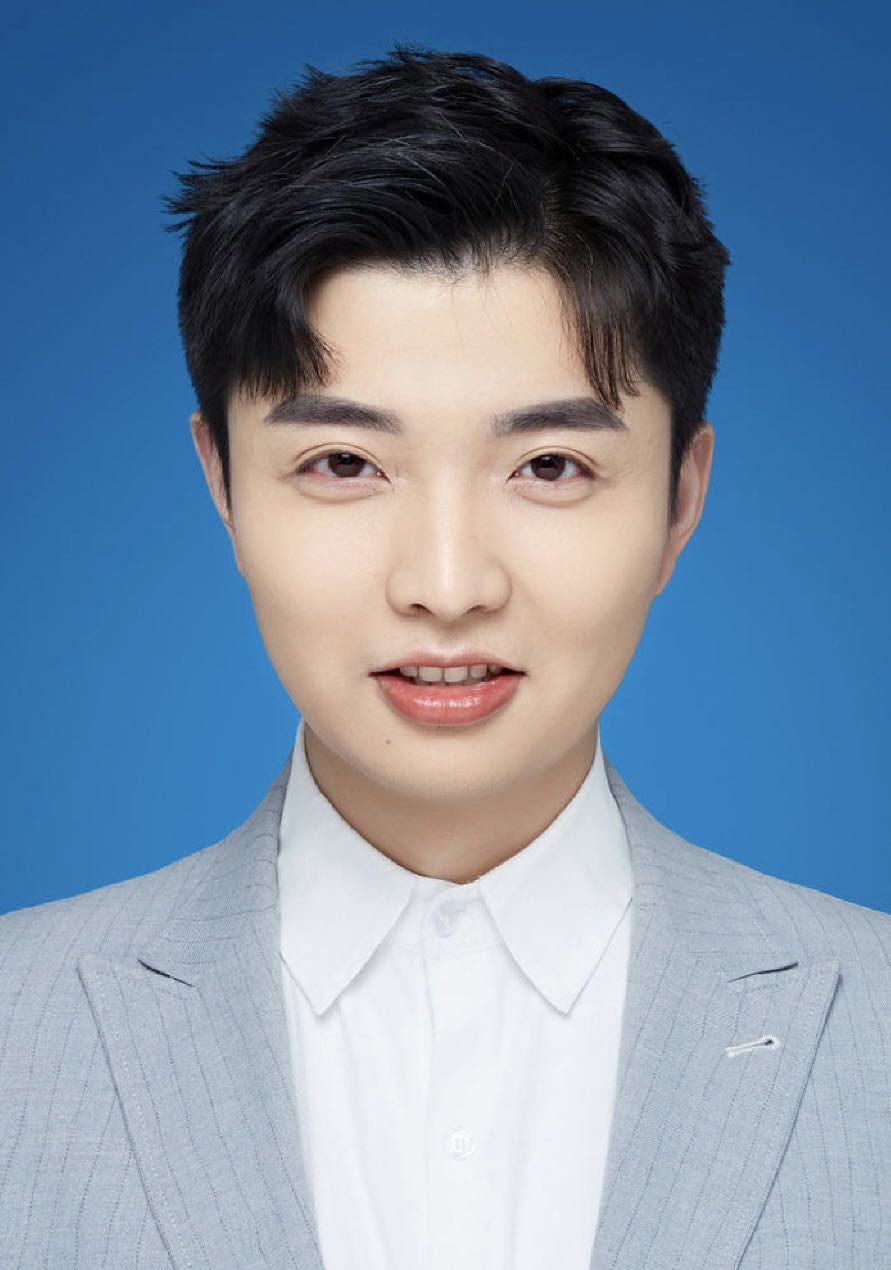}}]{Zixuan Li} received the B.E. degree from Anhui University in 2017, and the M.E. degree from the University of Chinese Academy of Sciences in 2021. He is a intern in the presidents office, Waytous Ltd., Beijing. 

His research interest cover computer vision ,communication engineering and autonomous driving.
\end{IEEEbiography}


\begin{IEEEbiography}[{\includegraphics[width=1in,height=1.25in,clip,keepaspectratio]{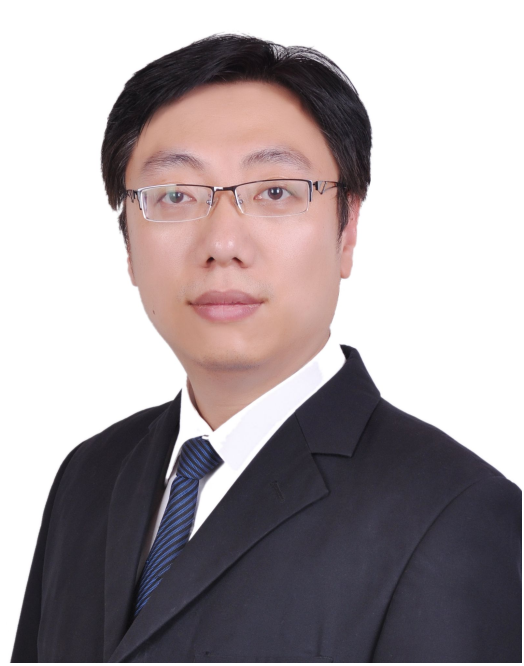}}]{Jinjun Wang} received the B.E. and M.E. degrees from Huazhong University of Science and Technology, China, in 2000 and 2003, respectively, and the Ph.D. degree from Nanyang Technological University, Singapore, in 2006. 

From 2006 to 2009, he was with NEC Laboratories America, Inc., as a Research Scientist, and from 2010 to 2013, he was with Epson Research and Development, Inc., as a Senior Research Scientist. He is currently a Professor with Xi’an Jiaotong University. His research interests include pattern classification, image/video enhancement and editing, content-based image/video annotation and retrieval, and semantic event detection.
\end{IEEEbiography}

\begin{IEEEbiography}[{\includegraphics[width=1in,height=1.25in,clip,keepaspectratio]{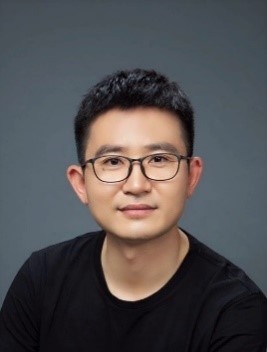}}]{Dongpu Cao} (Senior Member, IEEE) received the Ph.D. degree in mechanical engineering from Concordia University, Montreal, QC, Canada, in
2008.

He is a Professor with the school of mechanical engineering, Tsinghua University, Beijing, China. He has contributed more than 200 papers and three books. His current research focuses on driver cognition, automated driving, and cognitive autonomous driving.

Dr. Cao received the SAE Arch T. Colwell Merit Award in 2012, and three Best Paper Awards from the ASME and IEEE conferences. Dr. Cao serves as an Associate Editor for IEEE TRANSACTIONS ON VEHICULAR TECHNOLOGY, IEEE TRANSACTIONS ON INTELLIGENT TRANSPORTATION SYSTEMS, IEEE/ASME TRANSACTIONS ON MECHATRONICS, IEEE TRANSACTIONS ON INDUSTRIAL ELECTRONICS, IEEE/CAA JOURNAL OF AUTOMATICA SINICA and ASME JOURNAL OF DYNAMIC SYSTEMS, MEASUREMENT AND CONTROL. He was a Guest Editor for VEHICLE SYSTEM DYNAMICS and IEEE TRANSACTIONS ON SMC: SYSTEMS. He serves on the SAE Vehicle Dynamics Standards Committee and acts as the Co-Chair of IEEE ITSS Technical Committee on Cooperative Driving.
\end{IEEEbiography}

\begin{IEEEbiography}[{\includegraphics[width=1in,height=1.25in,clip,keepaspectratio]{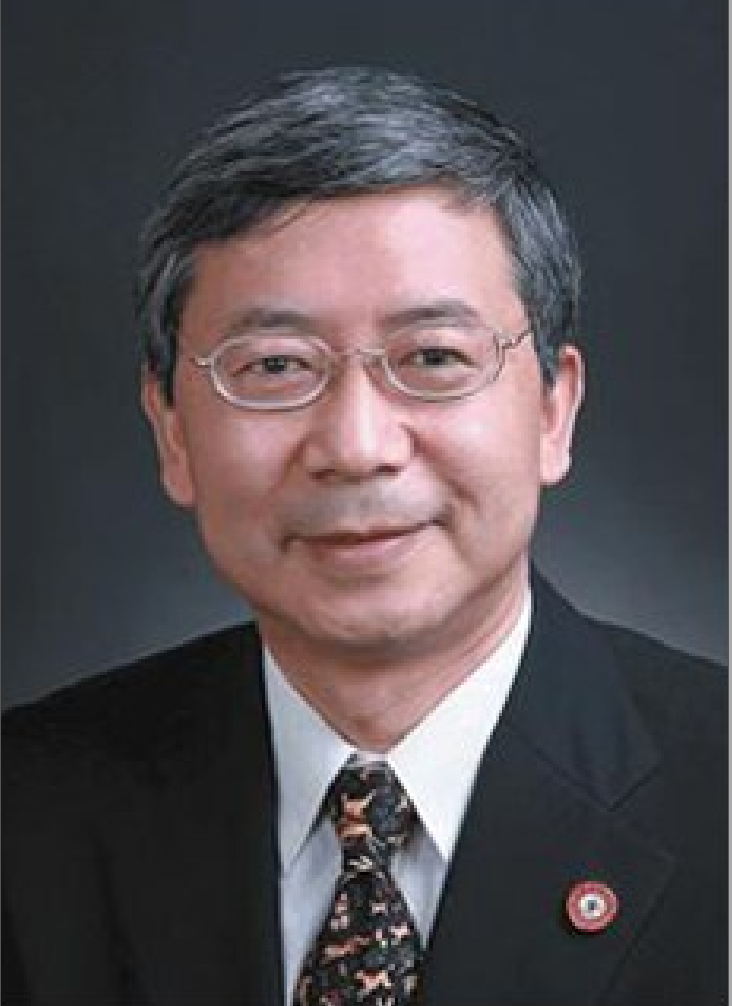}}]{Nanning Zheng} (Fellow, IEEE) graduated from the Department of Electrical Engineering, Xi'an Jiaotong University, Xi'an, China, in 1975. He received the M.S. degree in information and control engineering from Xi'an Jiaotong University in 1981 and the Ph.D. degree in electrical engineering from Keio University, Yokohama, Japan, in 1985. 

He jointed Xi'an Jiaotong University in 1975, where he is currently a Professor and the Director of the Institute of Artificial Intelligence and Robotics. His research interests include computer vision, pattern recognition and image processing, and hardware implementation of intelligent systems. 

Dr. Zheng became a member of the Chinese Academy of Engineering in 1999, and he is the Chinese Representative on the Governing Board of the International Association for Pattern Recognition.
\end{IEEEbiography}

\begin{IEEEbiography}[{\includegraphics[width=1in,height=1.25in,clip,keepaspectratio]{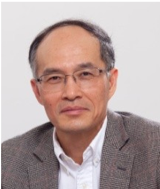}}]{Fei-Yue Wang} (Fellow, IEEE) received the Ph.D. degree in computer and systems engineering from the Rensselaer Polytechnic Institute, Troy, NY, USA, in 1990. 

He is currently a Professor and the Director of the State Key Laboratory of Intelligent Control and Management of Complex Systems, Institute of Automation, Chinese Academy of Sciences, Beijing, China. 

Prof. Wang was the Founding Editor-in-Chief of the INTERNATIONAL JOURNAL OF INTELLIGENT CONTROL AND SYSTEMS from 1995 to 2000, the Series on Intelligent Control and Intelligent Automation from 1996 to 2004, and the IEEE TRANSACTIONS ON INTELLIGENT TRANSPORTATION SYSTEMS. He was the Editor-in-Chief of the IEEE INTELLIGENT SYSTEMS from 2009 to 2011 and the IEEE TRANSACTIONS ON INTELLIGENT TRANSPORTATION SYSTEMS. He is the Editor-in-Chief of the IEEE/CAA JOURNAL OF AUTOMATIC SINICA. He is a member of Sigma Xi and an Elected Fellow of INCOSE, IFAC, ASME, and AAAS. He was the President of the IEEE Intelligent Transportation Systems Society from 2005 to 2007, the Chinese Association for Science and Technology, USA, in 2005, and the American Zhu Kezhen Education Foundation from 2007 to 2008. He is currently the Vice President and the Secretary General of the Chinese Association of Automation. 
\end{IEEEbiography}

\vfill

\end{document}